\documentclass{bmvc2k}

\title{FAR: A General Framework for Attributional Robustness}

\addauthor{Adam Ivankay}{aiv@zurich.ibm.com}{1}
\addauthor{Ivan Girardi}{ivg@zurich.ibm.com}{1}
\addauthor{Chiara Marchiori}{chi@zurich.ibm.com}{1}
\addauthor{Pascal Frossard}{pascal.frossard@epfl.ch}{2}

\addinstitution{
 IBM Research Zurich\\
R\"{u}schlikon, Switzerland
}
\addinstitution{
 \'{E}cole Polytechnique F\'{e}d\'{e}rale de Lausanne (EPFL)\\
 Lausanne, Switzerland
}

\runninghead{Ivankay et al.}{Framework for Attributional Robustness}

\usepackage{mathtools}
\usepackage{amsmath}
\usepackage{multirow}
\usepackage{array}
\usepackage{tikz}
\usepackage{pgfplots}
\pgfplotsset{compat=newest}
\usepackage{algorithm}
\usepackage{algorithmic}
\usepackage{booktabs}
\usepackage{wrapfig}
\usepackage{xcolor}
\usepackage[font=footnotesize]{caption}

\begin{document}

\maketitle

\begin{abstract}
Attribution maps are popular tools for explaining neural networks' predictions. By assigning an importance value to each input dimension that represents its impact towards the outcome, they give an intuitive explanation of the decision process. However, recent work has discovered vulnerability of these maps to imperceptible adversarial changes, which can prove critical in safety-relevant domains such as healthcare. Therefore, we define a novel generic \textit{framework for attributional robustness} (FAR) as general problem formulation for training models with robust attributions. This framework consist of a generic regularization term and training objective that minimize the maximal dissimilarity of attribution maps in a local neighbourhood of the input. We show that FAR is a generalized, less constrained formulation of currently existing training methods. We then propose two new instantiations of this framework, \textit{AAT} and \textit{AdvAAT}, that directly optimize for both robust attributions and predictions. Experiments performed on widely used vision datasets show that our methods perform better or comparably to current ones in terms of attributional robustness while being more generally applicable. We finally show that our methods mitigate undesired dependencies between attributional robustness and some training and estimation parameters, which seem to critically affect other competitor methods.
\end{abstract}

\section{Introduction}
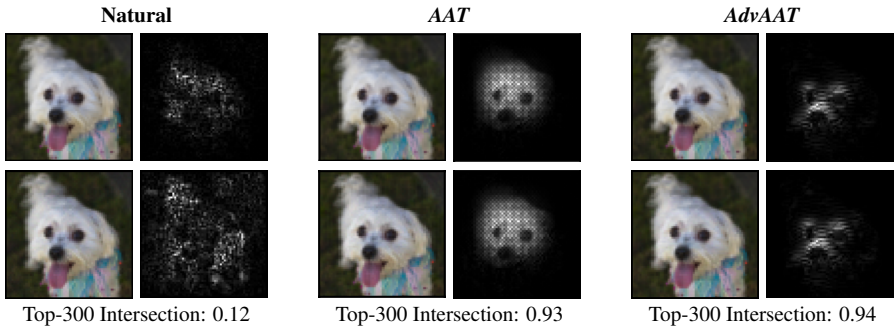
\begin{figure*}[t]
\footnotesize%
\centering
\newcommand{\sourceds}{example_rimagenet}
\newcommand{\imagewidth}{0.13}
\newcommand{\gridwidth}{0.16}
\newcommand{\gridheight}{0.14}
\newcommand{\figurewidth}{\textwidth}
\newcommand{\sideindent}{0.07}
\newcommand{\yoffset}{0.3}

\begin{tikzpicture}[x=\gridwidth\figurewidth, y=\gridheight\figurewidth]
\node[] (orig) at (0+\sideindent,1-\yoffset) {\includegraphics[width=\imagewidth\figurewidth]{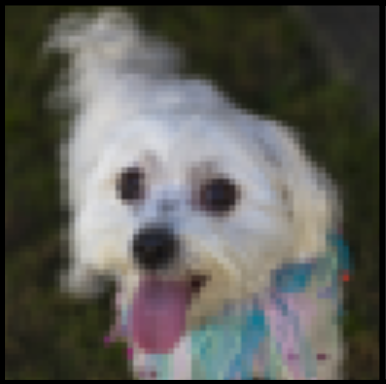}};
\node[] (orig) at (0+\sideindent,0-\yoffset) {\includegraphics[width=\imagewidth\figurewidth]{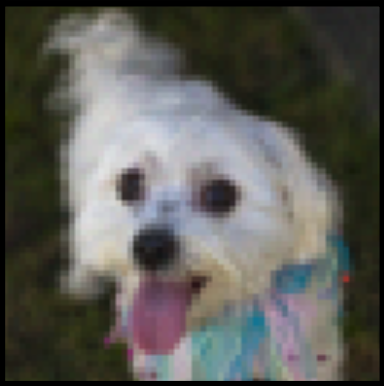}};
\node[] (orig) at (1-\sideindent,1-\yoffset) {\includegraphics[width=\imagewidth\figurewidth]{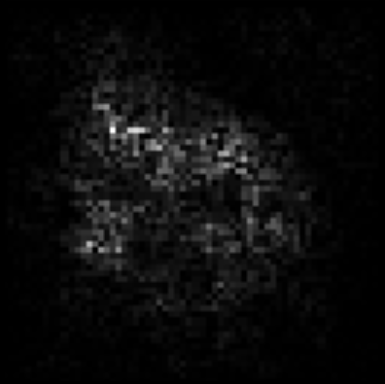}};
\node[] (orig) at (1-\sideindent,0-\yoffset) {\includegraphics[width=\imagewidth\figurewidth]{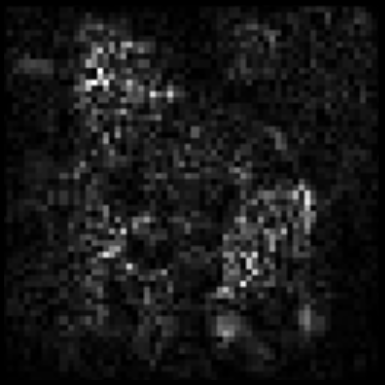}};
\node[] (natural) at (0.5,1.3) {\textbf{Natural}};
\node[] (natural) at (0.5,-0.9) {Top-300 Intersection: 0.12};

\node[] (orig) at (2+\sideindent,1-\yoffset) {\includegraphics[width=\imagewidth\figurewidth]{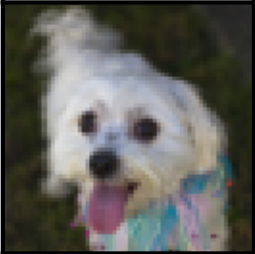}};
\node[] (orig) at (2+\sideindent,0-\yoffset) {\includegraphics[width=\imagewidth\figurewidth]{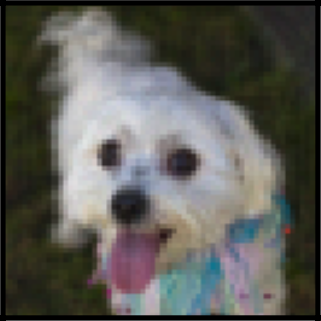}};
\node[] (orig) at (3-\sideindent,1-\yoffset) {\includegraphics[width=\imagewidth\figurewidth]{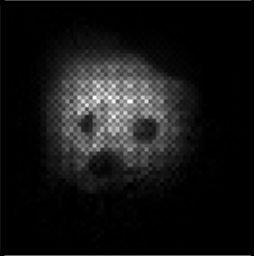}};
\node[] (orig) at (3-\sideindent,0-\yoffset) {\includegraphics[width=\imagewidth\figurewidth]{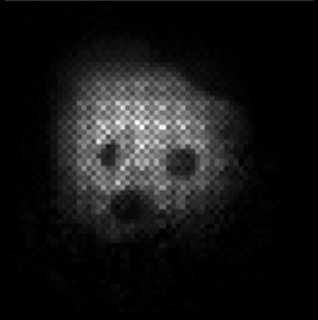}};
\node[] (natural) at (2.5,1.3) {\textbf{\textit{AAT}}};
\node[] (natural) at (2.5,-0.9) {Top-300 Intersection: 0.93};

\node[] (orig) at (4+\sideindent,1-\yoffset) {\includegraphics[width=\imagewidth\figurewidth]{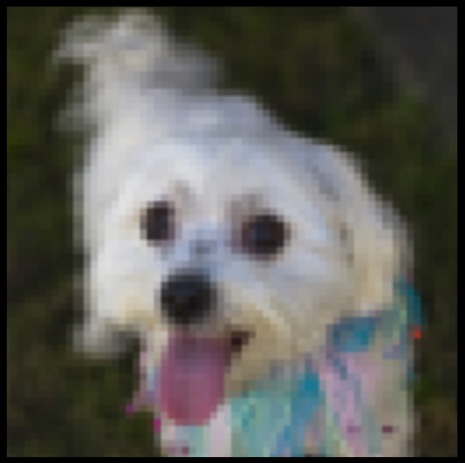}};
\node[] (orig) at (4+\sideindent,0-\yoffset) {\includegraphics[width=\imagewidth\figurewidth]{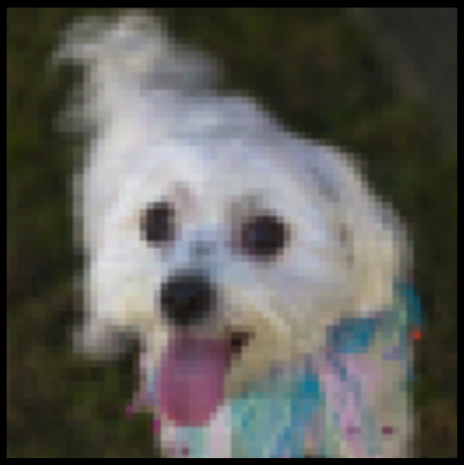}};
\node[] (orig) at (5-\sideindent,1-\yoffset) {\includegraphics[width=\imagewidth\figurewidth]{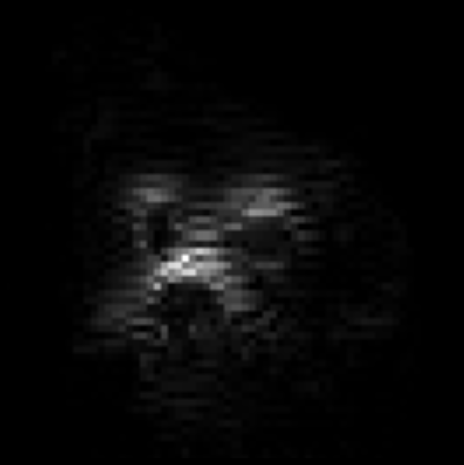}};
\node[] (orig) at (5-\sideindent,0-\yoffset) {\includegraphics[width=\imagewidth\figurewidth]{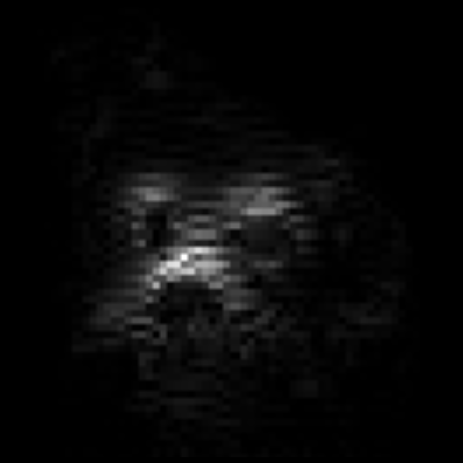}};
\node[] (natural) at (4.5,1.3) {\textbf{\textit{AdvAAT}}};
\node[] (natural) at (4.5,-0.9) {Top-300 Intersection: 0.94};

\end{tikzpicture}
\caption{Original and adversarial Integrated Gradients (IG) of the natural model (left), our \textit{AAT} (middle) and \textit{AdvAAT} (right) method on Restricted Imagenet. For each model, the upper row contains the unperturbed image in the left column and its IG saliency map in the right column. The lower row contains the corresponding perturbed image on the left and the resulting adversarial IG saliency map on the right. Our methods yield less noisy and more robust attribution maps (measured by the \textit{Top-300 intersection} of the highest attributed pixels of the unperturbed and perturbed image), while correctly classifying all images.}
\label{fig:example_fashionmnist}
\end{figure*}
As deep neural networks (DNNs) have become larger and deeper in recent years, their complexity also increased significantly. This makes it difficult for humans to understand and reason about their decision process. Therefore, methods that provide insight and reason about the prediction outcome of DNNs are crucial for successfully deploying these complex networks in real-life scenarios. \textit{Attribution maps} assign an importance value to each input dimension, which represents its \textit{influence} on the outcome of the decision. While these methods are fast to compute, they do not require specific domain knowledge to give interpretable explanations - \textit{the outcome is the given class because of the salient region in the input}, hence their popularity. Most of these methods, like Saliency \cite{saliencymap}, Integrated Gradients \cite{integratedgradients}, SmoothGrad \cite{smoothgrad} or DeepLIFT \cite{deeplift} utilize the input gradient of the network to construct the attribution map, giving explanation on the networks behaviour in a region around the given input.\\
{\indent}However, the authors in \cite{interpretationfragile} have successfully demonstrated that these explanations can be adversarially manipulated by adding carefully crafted perturbations to the input, fundamentally changing the attributions. This is  problematic in critical scenarios where the decision outcome needs to be accompanied by a sound explanation. Attackers could willingly manipulate the salient region in a digital pathology image, critically misleading the medical professional assessing the decision. In other instances, they could induce false bias in automatic credit scoring algorithms by pointing to the individuals race as reason for a low score. These are critical aspects that prevent DNNs from being adopted to solve real life problems and highlight the need for robust explanations. Figure \ref{fig:example_fashionmnist} exemplifies the adversarial fragility of attribution maps.\\
{\indent}In our work, we mitigate this fragility by making the following contributions:
\begin{itemize}
\itemsep0em
\item We define a \textit{framework for attributional robustness} (FAR) as general problem formulation for training robust attributions. Key aspects of this framework are:
    \begin{itemize}
        \itemsep0em
        \item It allows for separate optimization for robust predictions and explanations,
        \item It generalizes to more explanation methods and attribution distances than current methods,
        \item It allows for providing ground truth explanations.
    \end{itemize}
\item We provide two instantiations of FAR, our \textit{AAT} and \textit{AdvAAT} methods that directly optimize for maximal correlation between original and adversarial feature importance within a small $\mathrm{L}_\infty$ neighbourhood of the input. Experiments show that our methods outperform or perform similarly to other methods on widely-used vision datasets.
\item We identify undesired dependencies of gradient-based attribution maps on training and estimation parameters, which are mitigated by our methods.
\end{itemize}

\section{Preliminaries}
\subsection{Background}%
Let $f$ be a DNN classifier, $f_t(\mathbf{x})$ the output logit of $f$ for class $t$, $y {}={} \underset{t}{\arg\max}f_t(\mathbf{x})$ the predicted class and $g^\mathbf{x}_y(\mathbf{x})$ the input gradient for class $y$, given input $\mathbf{x}$. An \textit{\textbf{attribution method}} $\mathbf{\mathrm{S}}$ is a function that assigns a value to each element of $\mathbf{x}$ that represents its influence towards the prediction $y$ of DNN $f$.\\
{\indent}The simplest attribution map considered in this work is \textbf{\textit{Saliency Map}} (SM), defined as the element-wise absolute value of the input gradient, written as follows:
\begin{equation}
    \label{eqn:saliencymap}
    \mathrm{S}(\mathbf{x}, f) {}={} \mathrm{SM}(\mathbf{x}, f) {}={} |\nabla_{\mathbf{x}}\,f_y(\mathbf{x})|\,\coloneqq\,{g^y_\mathbf{x}(\mathbf{x})}^{\mathrm{abs}}
\end{equation}
{\indent}\textbf{\textit{Integrated Gradients}} (IG) is an axiomatic, smoothed version of SM \cite{integratedgradients}. IG is defined as the path integral from a predefined baseline $\mathbf{b}$ to the input $\mathbf{x}$ written in Equation (\ref{eqn:intgrad}).
\begin{equation}
\label{eqn:intgrad}
    \mathrm{S}(\mathbf{x}, f, \mathbf{b}) {}={} \mathrm{IG}(\mathbf{x}, f, \mathbf{b}) {}={} (\mathbf{x} - \mathbf{b}) \cdot \int_{\alpha=0}^{1} g^y_\mathbf{\tilde{x}}(\mathbf{\tilde{x}})|_{\mathbf{\tilde{x}} {}={} \mathbf{b} + \alpha(\mathbf{x} - \mathbf{b})}\, d\alpha
\end{equation}
where $\mathbf{b}$ can be chosen as an arbitrary input signal such as $\mathbf{0}$ or noise. It is a principled approach that fulfills sensitivity, implementation invariance and completeness axioms \cite{integratedgradients}, which makes it a commonly accepted and deployed explanation method.\\
{\indent}\textbf{\textit{Attributional Robustness}} (AR) refers to explanations' resistance towards adversarial perturbations. While there is no agreement of the definition of AR, most mathematical formulations build towards the conjecture that attribution maps should be similar for similar inputs. We define AR as written in Equation (\ref{eqn:ar}).
\begin{equation}
\label{eqn:ar}
    r(\mathrm{S}, \; \mathrm{S}^T) {}={} 1 {}-{} \max_{\mathbf{\tilde{x}}}\,\, d_{\mathrm{s}} \big[\, \mathrm{S}(\mathbf{\tilde{x}}, f), \; \mathrm{S}^T\big]
\end{equation}
given the constraints in the following Equation (\ref{eqn:arconstraints}),
\begin{equation}
\label{eqn:arconstraints}
    \|\mathbf{\tilde{x}}-\mathbf{x}\|_{\mathrm{p}}<\varepsilon \;\;\;\; \mathrm{and} \;\;\;\; \underset{t}{\arg\max} f_t(\mathbf{\tilde{x}}){}={}\underset{t}{\arg\max}f_t(\mathbf{x})
\end{equation}
where $r(\mathrm{S}, \; \mathrm{S}^T)$ denotes the attributional robustness of saliency map $\mathrm{S}$, $\mathrm{S}^T$ the target saliency map, $f$ the classifier, $\mathbf{\tilde{x}}$ and $\mathbf{x}$ the adversarial and original inputs, $\varepsilon$ a small bound on the $\mathrm{L_p}$-norm of the input change and $d_{\mathrm{s}}$ a (scaled) dissimilarity metric for attribution maps.%
\subsection{Related work}%
\label{sec:relatedwork}%
{\indent}First efforts for mitigating this fragility were introduced in \cite{alignmentconnection}. The authors proved theoretical connections between adversarial robustness and the \textit{alignment} ${\langle}g^y_\mathbf{x}(\mathbf{x}), \mathbf{x}\rangle$ of prediction input gradients $g^y_\mathbf{x}(\mathbf{x})$ and input $\mathbf{x}$. Based on this consideration, it has been shown that adversarially trained networks (in their predictions) have increased attributional robustness \cite{igsum,alignment,smoothedgeometry}. On the other hand, the authors of \cite{alignment} have achieved state of the art AR by maximizing alignment with a regularization term during training. However, the input-gradient alignment as an upper bound for robustness strictly holds only in a local linear neighbourhood of the input. Moreover, it depends on both the input and its gradients, as it computes the inner product between them. This is a strong assumption, as zero-input areas do not contribute to the alignment, no matter their gradient.\\%
{\indent}An axiomatic approach to achieve robust saliency maps has been introduced in \cite{igsum}. The authors use Integrated Gradients (IG) \cite{integratedgradients} and its theoretical properties to attain robust attributions. In other research, the works of \cite{geometryblame} and \cite{smoothedgeometry} analyzed the local geometry of the classifier, effectively reducing attribution fragility by smoothing the curvature of the classifier. This is consistent with previous observations of correlations between adversarially robust networks and robust attributions \cite{alignmentconnection,alignment}.\\
{\indent}While there has been significant effort in mitigating adversarial vulnerability of attribution maps, most works have crucial shortcomings. First, they jointly optimize for both adversarial and attributional robustness. This does not allow for a separate analysis of the two notions. Second, they achieve AR by regularizing curvature or gradients, therefore can not be straightforwardly defined for explanations that are not based on gradients. Third, none support defining target explanations, which is useful for scenarios described in \cite{rightreasons} or \cite{reflectivenet}. Lastly, the method from \cite{alignment} computes the inner product between input and gradients, thus coupling input data and gradient domains. This can not be straightforwardly defined for non-continuous inputs like categorical variables, text or other constraint inputs in multimodal problems.%

\section{Framework for Attributional Robustness}
In this section, we introduce FAR, our general problem formulation of solving AR in DNNs. It consists of two generic training objectives for robust attributions and predictions. Then, we derive existing robust attribution training methods from these objectives, showcasing the general nature of our formulation.\\
{\indent}In order to mitigate the shortcomings of current attributional robustness methods, we extend the classical notion of adversarial training \cite{madryadversarial} to attribution maps. Thus, we make the following considerations:
\begin{itemize}
    \itemsep0em
    \item \textit{Similar attribution maps.} The main assumption of our framework is that similar inputs should give near-identical explanations, as defined in Equation (\ref{eqn:ar}). It is often argued that not all input features matter towards the prediction, but only a certain subset, or even only their relative ranks \cite{interpretationfragile}. Therefore, we measure attribution similarity with metrics that reflect these considerations, like the \textit{Kendall's rank order correlation} (\texttt{CO}) \cite{kendall} or the \textit{Top-K intersection} (\texttt{IN}) \cite{interpretationfragile}, as done in most of the related work.
    \item \textit{Perceptually identical inputs} and \textit{unchanged prediction outcome.} Analogously to traditional adversarial training, we utilize widely-used $\mathrm{L}_\mathrm{p}$-ball restrictions of size $\varepsilon$ around the input to ensure unchanged ground truth labels of the data. Moreover, we require the same predicted class $\underset{t}{\arg\max}f_t(\mathbf{x}){}={}\underset{t}{\arg\max}f_t(\mathbf{x}_\mathrm{adv})$ of original and perturbed inputs. The latter constraint motivates our assumption that similar inputs should have similar explanations.
    \item \textit{Target attributions} and \textit{identical prediction outcomes.} Generally, ground truth for attribution maps is not available, therefore, we use the attributions of the unperturbed inputs as targets. However, allowing to provide these targets could provide useful for datasets in which they are given.
\end{itemize}
\subsection{Optimization Problem}
\label{sec:far}
Given the above points, we define our framework as a \textit{regularization term} added to the classification loss and a \textit{robust training loss}, which formulate generic objectives for robustifying any smooth attribution method and dissimilarity. The regularization term is used to enhance robustness of the attributions separately from the inference outcome, while the robust training loss jointly encourages adversarial and attributional robustness. Thus, the training objectives of the framework become the following \textit{min-max} optimization problems written in Equations (\ref{eqn:robustregularizationtraining}) and (\ref{eqn:robusttraining}).
\begin{equation}
\label{eqn:robustregularizationtraining}
\begin{split}
    \theta^* {}={} \underset{\theta}{\arg\min} \,  \sum_{\mathbf{x} \in \mathrm{D}} \, \big\{\, l(\mathbf{x}, y, f) \,
    + \lambda \, \cdot \, \max_{\|\mathbf{\tilde x} - \mathbf{x}\|_{\mathrm{p}} < \varepsilon} \, d_{\mathrm{s}} \big[\, \mathrm{S}(\mathbf{\tilde{x}}, f), \mathrm{S}^T(\mathbf{x}, f)\, \big] \; \big\}
\end{split}
\end{equation}
or
\begin{equation}
\label{eqn:robusttraining}
\begin{split}
    \theta^* {}={} \underset{\theta}{\arg\min} \,  \sum_{\mathbf{x} \in \mathrm{D}} \,  \max_{\|\mathbf{\tilde x} - \mathbf{x}\|_{\mathrm{p}} < \varepsilon} \, \big\{\, l(\mathbf{\tilde x}, y, f)
    + \lambda \, \cdot \, d_{\mathrm{s}} \big[\, \mathrm{S}(\mathbf{\tilde{x}}, f), \mathrm{S}^T(\mathbf{x}, f)\, \big] \, \big\}
\end{split}
\end{equation}
where $f$ denotes the classifier, $l$ the classification loss, $d_{\mathrm{s}}$ is any smooth dissimilarity metric between the saliency map $\mathrm{S}$ and a target saliency map $\mathrm{S}^T$, $\mathrm{p}$ denotes an $\varepsilon$-bounded norm base, $y$ the target class and $\theta^*$ the optimal parameters of the classifier $f$ trained on dataset $\mathrm{D}$. $\lambda$ controls the robust attribution regularization. Note that while Equation (\ref{eqn:robustregularizationtraining}) allows for solely optimizing for robust attributions, the training loss described in Equation (\ref{eqn:robusttraining}) encourages both robust attributions \textit{and} predictions. The inner maximizations of Equations (\ref{eqn:robustregularizationtraining}) and (\ref{eqn:robusttraining}) are solved via the IFIA (Algorithm \ref{alg:ifia}) and the Adversarial IFIA (Algorithm \ref{alg:advifia}) algorithms respectively, written in Figure \ref{fig:innermaximizations}.\\
{\indent}Formulating the AR problem as above has the following advantages. First, the choice of $\mathrm{S}$ is not fixed - the framework can be used to robustify any saliency map. Second, the domain of explanations and input data is not coupled, hence the shortcomings of current methods described in Section \ref{sec:relatedwork} do not exist for our framework. Third, the choice of $\mathrm{S}^T$ is not fixed, therefore target (ground truth) explanations can be provided if present, and robust explanations can be trained with respect to these. Fourth, the dissimilarity metric $d_{\mathrm{s}}$ can be chosen to any smooth $d_{\mathrm{s}}$, depending on the use case. Lastly, by varying the regularization parameters, we can adjust the trade-off between robust attributions and predictions.%
%
%
\newcommand{\algifia}[1]{%
  \begin{minipage}[t]{.44\linewidth}
    \begin{algorithm}[H]
\small
\caption{IFIA}\label{alg:ifia}
\textbf{Input}: Classifier $f$, input $\mathbf{x}$, target class $t$, attribution map $\mathrm{S}$, dissimilarity metric $d_{\mathrm{s}}$, norm $\mathrm{p}$ and bound $\varepsilon$, step size $\eta$, iterations $N$, data input bounds $b$\\
\textbf{Output}: Adversarial example $\mathbf{x_\mathrm{adv}}$%
\begin{algorithmic}[1]
\STATE $\mathbf{x_\mathrm{adv}} \gets \mathbf{x}$
\WHILE {$i \leq N$ \AND $\underset{t}{\arg\max}f_t(\mathbf{x_\mathrm{adv}}) {}={} \underset{t}{\arg\max}f_t(\mathbf{x})$}
    \STATE $\mathbf{g}_t \gets \nabla_{\mathbf{x_\mathrm{adv}}}\, d_{\mathrm{s}} \big[\, \mathrm{S}(\mathbf{x_\mathrm{adv}}, f),\; \mathrm{S}(\mathbf{x}, f)\, \big]$
    \STATE $\mathbf{x_\mathrm{adv}} \gets \mathbf{x_\mathrm{adv}} + \eta\, \cdot \mathtt{Normalize}_{\mathrm{p}}\big(\, \mathbf{g}_t \, \big)$
    \STATE $\mathbf{x_\mathrm{adv}} \gets \mathtt{Project}_{\mathrm{p}}\big(\, \mathbf{x_\mathrm{adv}}, \mathbf{x}, \varepsilon, b\, \big)$
\ENDWHILE
\end{algorithmic}
\end{algorithm}%
  \end{minipage}%
}
\newcommand{\algadvifia}[1]{%
  \begin{minipage}[t]{.42\linewidth}
    \begin{algorithm}[H]
\small
\caption{Adversarial IFIA}\label{alg:advifia}
\textbf{Input}: Classifier $f$, input $\mathbf{x}$, target class $t$, classification loss $l$, attribution map $\mathrm{S}$, dissimilarity metric $d_{\mathrm{s}}$, norm $\mathrm{p}$ and bound $\varepsilon$, step size $\eta$, iterations $N$, data input bounds $b$\\
\textbf{Output}: Adversarial example $\mathbf{x_\mathrm{adv}}$%
\begin{algorithmic}[1]
\STATE $\mathbf{x_\mathrm{adv}} \gets \mathbf{x}$
\WHILE {$i \leq N$}
    \STATE $\mathbf{g}_t \gets \nabla_{\mathbf{x_\mathrm{adv}}}\, \{l(\mathbf{x_\mathrm{adv}}, t, f) + \lambda \, \cdot \,  d_{\mathrm{s}} \big[\, \mathrm{S}(\mathbf{x_\mathrm{adv}}, f),\; \mathrm{S}(\mathbf{x}, f)\, \big] \}$
    \STATE $\mathbf{x_\mathrm{adv}} \gets \mathbf{x_\mathrm{adv}} + \eta\, \cdot \mathtt{Normalize}_{\mathrm{p}}\big(\, \mathbf{g}_t \, \big)$
    \STATE $\mathbf{x_\mathrm{adv}} \gets \mathtt{Project}_{\mathrm{p}}\big(\, \mathbf{x_\mathrm{adv}}, \mathbf{x}, \varepsilon, b\, \big)$
\ENDWHILE
\end{algorithmic}
\end{algorithm}%
  \end{minipage}%
}
\begin{figure}[h]
  \null\hfill \algifia{alg1} \hfill \algadvifia{alg2} \hfill\null\par
  \caption{The IFIA (left) and Adversarial IFIA (right) attacks used to solve the inner maximizations of our framework. IFIA is also used to estimate AR during evaluation.}
  \label{fig:innermaximizations}
\end{figure}
\subsection{Recovering existing objectives}
In this section, we show that our formulation of the robustness optimization problem is a generalization of already existing methods to train robust attribution maps.\\
{\indent}\textit{\textbf{Madry's Robust Prediction}} \cite{madryadversarial} can be recovered by utilizing the training objective in Equation (\ref{eqn:robusttraining}) with $\lambda {}={} 0$. It becomes as follows:
\begin{equation}
\label{eqn:madryrobust}
    \theta^* {}={} \underset{\theta}{\arg\min} \,  \sum_{\mathbf{x} \in \mathrm{D}} \, \max_{\|\mathbf{\tilde x} - \mathbf{x}\|_{\mathrm{p}} < \varepsilon} \, l(\mathbf{\tilde x}, y, f)
\end{equation}
{\indent}The \textit{\textbf{Axiomatic Attribution Regularization}} terms (IG-NORM and IG-SUM-NORM) in \cite{igsum} can be recovered using the regularization term in Equation (\ref{eqn:robustregularizationtraining}) with the IG attribution map $\mathrm{S} {}={} \mathrm{IG}(\mathbf{\tilde{x}}, \mathbf{x})$, where the baseline of IG is set to $\mathbf{b} {}={} \mathbf{x}$ and the dissimilarity function to $d_{\mathrm{s}}(\mathbf{x}, \mathbf{y}) {}={}  \|\mathbf{x} {}-{} \mathbf{y}\|_1$. As such, Equation (\ref{eqn:robustregularizationtraining}) becomes as follows.
\begin{equation}
\label{eqn:ignormequation}
\begin{split}
    \theta^* {}={} \underset{\theta}{\arg\min} \,  \sum_{\mathbf{x} \in \mathrm{D}} \, \big\{\, l(\mathbf{x}, y, f)
    + \lambda \, \cdot \, \max_{\|\mathbf{\tilde x} - \mathbf{x}\|_{\mathrm{p}} < \varepsilon} \, \|\mathrm{IG}(\mathbf{\tilde{x}}, \mathbf{x})\,\|_1 \big\}
\end{split}
\end{equation}
The IG-SUM-NORM training objective in \cite{igsum} can be analogously derived from Equation (\ref{eqn:robusttraining}). Note that $\mathrm{IG}(\mathbf{x}, \; \mathbf{x}) {}={} 0$ holds due to the completeness axiom of IG \cite{igsum}.\\
{\indent}The input-gradient \textbf{\textit{Spatial Alignment}} regularization term introduced in \cite{alignment} corresponds to utilizing the sum of positive spatial alignment of true class input gradients and the negative spatial alignment of the second largest logits input gradient as attribution map $\mathrm{S}$, written in the following Equation (\ref{eqn:alignmentobjective}):
\begin{equation}
\label{eqn:alignmentobjective}
\begin{split}
    \mathrm{S}(\mathbf{x}, f, y, \bar{y}) {}={} \mathrm{cos} \big[ {g^y_\mathbf{x}(\mathbf{x})}, \mathbf{x} \big] \,
    -{} \, \mathrm{cos} \big[ {g^{\bar{y}}_\mathbf{x}(\mathbf{x})}, \mathbf{x} \big]
\end{split}
\end{equation}
$\mathrm{cos}$ denotes the \textit{pointwise} cosine similarity between the input gradient and the image. $\bar{y}$ is the second largest class' logit, the rest of the notation is kept from previous sections. By using $d_{\mathrm{s}}(\mathbf{x}, \mathbf{y}) {}={} \log \, \big\{ 1 + \exp \, \big[-{} \sum_{i\in\mathrm{dim}(\mathbf{x})}(x_i {}-{} y_i) \, \big] \, \big\}$ and omitting the use of target saliency maps, the regularization term in \cite{alignment} can be recovered from Equation (\ref{eqn:robustregularizationtraining}). See the technical appendix for a more detailed proof of these equations.

\section{Adversarial Attributional Training}
Next, we introduce \textit{AAT} and \textit{AdvAAT}, our instantiations of FAR that achieve robust attribution maps through optimizing for maximal correlation of explanations within a small local $\mathrm{L}_{\infty}$ neighbourhood of the input.\\
{\indent}Using our framework, we formalize the \textbf{\textit{adversarial attributional training}} objectives, consisting of a \textit{regularization term} (\textit{AAT}) that optimizes directly for robust attributions, and a \textit{robust training loss} (\textit{AdvAAT}) used to achieve both robust predictions and attributions. We choose the \textit{Pearson correlation coefficient} \cite{pcc} as attribution similarity, as it is a good proxy for optimizing for discrete rank correlations like \texttt{CO} and \texttt{IN}. These cannot be used directly due to their non-differentiable nature. We choose the aforementioned IG as attribution map, as it is a widely accepted axiomatic explanation method. Our attribution targets are the saliency maps of the unperturbed inputs. This leads to the following optimization regularization term (\ref{eqn:attradvregularization}) and loss (\ref{eqn:attradvobjective}) respectively.
\begin{equation}
\label{eqn:attradvregularization}
\begin{split}
    \theta_{\mathrm{AAT}}^* {}={} \underset{\theta}{\arg\min} \,  \sum_{\mathbf{x} \in \mathrm{D}} \, \big\{\, l(\mathbf{x}, y, f)
    + \lambda \, \cdot \, \max_{\|\mathbf{\tilde x} - \mathbf{x}\|_{\infty} < \varepsilon} \, \mathrm{PCL} \big[\, \,\mathrm{IG}(\mathbf{\tilde{x}}, \mathbf{0}), \;\mathrm{IG}(\mathbf{x}, \mathbf{0})\, \big]\, \big\}
\end{split}
\end{equation}
\begin{equation}
\label{eqn:attradvobjective}
\begin{split}
    \theta_{\mathrm{AdvAAT}}^* {}={} \underset{\theta}{\arg\min} \,  \sum_{\mathbf{x} \in \mathrm{D}} \, \max_{\|\mathbf{\tilde x} - \mathbf{x}\|_{\infty} < \varepsilon} \, \big\{\, l(\mathbf{\tilde x}, y, f)
    + \lambda \, \cdot \, \mathrm{PCL} \big[\, \,\mathrm{IG}(\mathbf{\tilde{x}}, \mathbf{0}), \;\mathrm{IG}(\mathbf{x}, \mathbf{0})\, \big] \, \big\}
\end{split}
\end{equation}
\begin{wraptable}{R}{0.65\textwidth}%
\begin{tabular}{llrrrr}
\toprule
\shortstack{\texttt{Data}}            & \shortstack{\texttt{Model}} & \shortstack{\texttt{NA}\\\texttt{(\%)}} & \shortstack{\texttt{AA}\\\texttt{(\%)}} & \shortstack{\texttt{IN}} & \shortstack{\texttt{CO}}      \\ \hline\hline
\multirow{6}{*}{MNIST}        & Nat            & 99.4      & 12.1       & 0.43        & 0.10        \\ 
                              & Adv            & 98.9      & 92.7      & 0.52        & 0.19        \\ 
                              & Align          & 98.7      & 2.6       & 0.52        & 0.40        \\ 
                              & Align (s.)          & 95.2      & 12.3       & 0.58        & 0.43        \\ 
                              & IG-SN           & *98.3     & *88.2     & *0.72       & *0.31       \\ 
                              & \textit{AAT}           & 98.4      & 0.0       & 0.76        & 0.72       \\ 
                              & \textit{AdvAAT}         & 98.7      & 77.1      & \textbf{0.77}        & \textbf{0.73}        \\
\hline
\multirow{6}{*}{\shortstack{Fashion-\\-MNIST}}& Nat            & 91.5      & 11.0       & 0.43        & 0.20        \\ 
                              & Adv            & 87.1      & 69.9      & 071        & 0.58        \\ 
                              & Align          & 90.2      & 30.5      & 0.48        & 0.60        \\ 
                              & Align (s.)         & 85.4      & 20.3      & 0.50        & 0.45        \\ 
                              & IG-SN           & *85.4     & *70.3     & *0.72       & *0.67       \\ 
                              & \textit{AAT}           & 89.8      & 0.01      & 0.80        & \textbf{0.82}        \\ 
                              & \textit{AdvAAT}         & 86.7      & 41.4      & \textbf{0.81}       & \textbf{0.82}        \\
\hline
\multirow{6}{*}{\shortstack{CIFAR-\\-10}}        & Nat            & 89.9      & 0.0       & 0.17        & -0.02        \\ 
                              & Adv            & 80.3      & 43.9      & 0.66        & 0.66        \\ 
                              & Align          & *89.8     & *37.6     & *\textbf{0.93}       & *\textbf{0.92}       \\ 
                              & IG-SN           & -           & -           & -           & -           \\ 
                              & \textit{AAT}           & 73.7      & 0.4       & 0.86        & 0.70        \\ 
                              & \textit{AdvAAT}         & 72.2      & 24.9      & 0.90        & 0.71        \\
\hline
\multirow{6}{*}{GTSRB}        & Nat            & 98.5      & 14.7      & 0.39        & 0.19        \\ 
                              & Adv            & 94.9     & 66.7     & 0.72       & 0.64       \\ 
                              & Align          & *98.5     & *84.7     & *\textbf{0.92}       & *\textbf{0.89}       \\ 
                              & IG-SN           & *95.7     & *77.1     & *0.74       & *0.77       \\ 
                              & \textit{AAT}           & 95.6      & 26.9      & 0.75        & 0.79        \\ 
                              & \textit{AdvAAT}         & 91.7      & 65.9      & 0.84        & 0.80        \\
\hline
\multirow{6}{*}{\shortstack{Restr.\\Imagenet}}        & Nat            & 89.1      & 0.0      & 0.08        & 0.20        \\ 
                              & Adv            & 80.0     & 68.2     & 0.81       & 0.78       \\ 
                              & Align          & 82.3     & 67.7     & \textbf{0.92}       & \textbf{0.86}       \\ 
                              & IG-SN           & -     & -     & -       & -       \\ 
                              & \textit{AAT}           & 88.4      & 0.02      & 0.91        & 0.78        \\ 
                              & \textit{AdvAAT}         & 80.2      & 61.1      & 0.90        & 0.79        \\
\hline
\end{tabular}
\caption{Estimated attributional robustness (\textit{Top-K intersection} \texttt{IN} and \textit{Kendall's rank order correlation} \texttt{CO}) of the models trained naturally (Nat), adversarially (Adv), \textit{alignment}-based (Align), IG-SUM-NORM-based (IG-SN) as well as with our \textit{AAT} and \textit{AdvAAT} objectives. Their natural and adversarial accuracy is given in the \texttt{NA} and \texttt{AA} columns. Numbers indicated with an asterix (*) are taken from the respective work and not reproduced by us. Align (s.) denotes the \textit{alignment}-based method with input images scaled between -1 and 1.}\label{tbl:robustnessresults}
\end{wraptable}%
with $\mathrm{PCL} {}={} 1 {}-{} \frac{\mathrm{PCC} + 1}{2}$ denoting the loss derived from \textit{Pearson correlation coefficient} $\mathrm{PCC}$, the rest of the notation is kept from the previous sections.\\%
{\indent}The outer minimization is solved by standard gradient descent of the loss in the network parameter space. The inner maximizations of our AAT and AdvAAT methods during training are solved with the IFIA (Algorithm \ref{alg:ifia}) and Adversarial IFIA (Algorithm \ref{alg:advifia}) attacks from the previous section. During these attacks, we only approximate the second derivative of the ReLU networks with the second derivative of the Softplus activation  $\nabla^2_\mathbf{x}\,\mathrm{ReLU}(\mathbf{x}) {}={} \beta \cdot \mathrm{sigmoid}(\beta \cdot \mathbf{x}) \cdot \big[ \,1 -  \mathrm{sigmoid}(\beta \cdot \mathbf{x})\, \big]$, where $\mathrm{sigmoid}(\mathbf{x}) = \frac{1}{1 + e^{-\mathbf{x}}}$ and $\beta$ controls the approximation tightness of the ReLU \cite{geometryblame}. As such, we decouple the attribution maps from the actual estimation of their robustness.

\section{Experiments and Results}
\label{sec:experiments}
In this section, we report the experimental setup and evaluation of our \textit{AAT} and \textit{AdvAAT} methods. Utilizing the datasets MNIST \cite{mnist}, Fashion-MNIST \cite{fashionmnist}, CIFAR-10 \cite{cifar}, GTSRB \cite{gtsrb} and Restricted Imagenet \cite{restrictedimagenet}, we show that our methods outperform current state of the art on the former two datasets and perform comparably to state of the art on the latter three in terms of attributional robustness. Additionally, to our knowledge, we are the first to experimentally show the dependency of attribution robustness on the weight initialization of the networks and argue that training with our objectives lessens these dependencies. Moreover, we show that the tightness parameter $\beta$ in the approximation of second order ReLU gradient significantly influences the robustness estimation.%
%
\paragraph{Setup.} We compare three state of the art attribution robustification methods (Adv, IG-SN and Align), taken from \cite{madryadversarial}, \cite{igsum} and \cite{alignment} respectively, and a naturally trained (Nat) models' attributional robustness to networks trained with our robust training objectives from Equations (\ref{eqn:robustregularizationtraining}) and (\ref{eqn:robusttraining}) (\textit{AAT} and \textit{AdvAAT}), on the aforementioned five datasets. For MNIST and Fashion-MNIST, we train a two-layer convolutional neural network, for the other datasets we use a ResNet taken from \cite{resnet}. In order to evaluate the attributional robustness of each model, the IFIA attack from \cite{interpretationfragile} (Algorithm \ref{alg:ifia}) is used, utilizing the proposed \textit{Top-K intersection} attack from \cite{interpretationfragile}. We use the IG attribution map and report the \textit{Top-K intersection} (\texttt{IN}) of original and adversarial attribution map as well as their \textit{Kendall rank order correlation} (\texttt{CO}) as robustness metrics. The natural and adversarial accuracy (\texttt{NA} and \texttt{AA}) of the models are also reported. \texttt{AA} is estimated with the PGD attack from the authors of \cite{madryadversarial}. A detailed description of the architectures, training and evaluation details can be found in the supplemental appendix. Table \ref{tbl:robustnessresults} contains the results of the comparison experiments. The results are run three times with different data splits and random seeds, and the average results are given.%
%
%
\paragraph{Numerical analysis.} Based on Table \ref{tbl:robustnessresults}, we make the following conclusions. First, our methods outperform all other state of the art methods on MNIST and Fashion-MNIST. On the datasets CIFAR-10, GTSRB and Restricted Imagenet, our methods perform comparably to state of the art in terms of \texttt{IN}, while giving slightly worse results in terms of \texttt{CO}. Hence, we conclude that while \textit{AAT} and \textit{AdvAAT} do not outperform Align, they give promising results while being more general and wider applicable, as described in Section \ref{sec:far}. This is backed by the phenomenon that our methods perform significantly better on MNIST and Fashion-MNIST than Align. We argue that this is due to Align being dependant on the nature of the data. A large proportion of the data are black pixels. Along these dimensions, the alignment from Equation (\ref{eqn:alignmentobjective}) is inherently zero, independently of the gradients. Therefore, Align does not provide an optimization target along these dimensions. Moreover, white pixels are targeted to have large gradients (in alignment terms), but gradient saturation leads to small gradients for these pixels, further worsening optimization with Align on the two MNIST datasets. Our methods do not suffer from these shortcomings, as they provide optimization targets for each input dimension, independently of their values. We evaluated Align on the MNIST datasets with an input scaling between [-1, 1] as well, indicated as Align (s.) in Table \ref{tbl:robustnessresults}. However, we see almost no improvement in terms of \texttt{IN} and \texttt{CO} compared to scaling between [0, 1] (Align). We believe that this is due to the arbitrary choice of input bounds. A lower bound of -1 encourages negative gradients, another arbitrary valid lower bound of 0.2 would encourage positive ones in the same dimensions. This highlights the flaws of the alignment-based method even more, namely that targets are not input shift invariant.\\%
Our second conclusion comes from comparing our \textit{AAT} method to \textit{AdvAAT}. \textit{AAT} achieves slightly worse attribution robustness than \textit{AdvAAT}, but significantly worse adversarial accuracy for all datasets experimented on. This leads us to believe that while adversarial robustness does increase attributional robustness, the reverse is only limitedly true. We leave the theoretical analysis of this phenomenon to future work.%
\newcommand{\pltwidth}{0.5}
\newcommand{\lambdaaat}{%
  \begin{minipage}[t]{.43\linewidth}
\begin{tikzpicture}
\begin{axis}[
xmin = 0.0, xmax = 1.5,
ymin = 0.0, ymax = 1.0,
grid = both,
minor tick num = 1,
major grid style = {lightgray!25},
minor grid style = {lightgray!25},
width = \linewidth,
height = 0.6\linewidth,
xlabel = {$\lambda$},
title=\textbf{\textit{AAT}},
every axis title/.style={above,at={(0.5,1.1)}}
]
\addplot[smooth, black, opacity=0.6] table[x={x}, y={na}] {data/lambda_aat.dat};
\addplot[smooth, orange, opacity=0.6] table[x={x}, y={aa}] {data/lambda_aat.dat};
\addplot[smooth, thick, blue] table[x={x}, y={in}] {data/lambda_aat.dat};
\addplot[smooth, thick, red] table[x={x}, y={co}] {data/lambda_aat.dat};

\end{axis}
\end{tikzpicture}

  \end{minipage}%
}%
\newcommand{\lambdaadvaat}{%
  \begin{minipage}[t]{.43\linewidth}
\begin{tikzpicture}
\begin{axis}[
xmin = 0.0, xmax = 1.5,
ymin = 0.0, ymax = 1.0,
grid = both,
minor tick num = 1,
major grid style = {lightgray!25},
minor grid style = {lightgray!25},
width = \linewidth,
height = 0.6\linewidth,
xlabel = {$\lambda$},
title=\textbf{\textit{AdvAAT}},
every axis title/.style={above,at={(0.5,1.1)}}
]
\addplot[smooth, black, opacity=0.6] table[x={x}, y={na}] {data/lambda_advaat.dat};
\addplot[smooth, orange, opacity=0.6] table[x={x}, y={aa}] {data/lambda_advaat.dat};
\addplot[smooth, thick, blue] table[x={x}, y={in}] {data/lambda_advaat.dat};
\addplot[smooth, thick, red] table[x={x}, y={co}] {data/lambda_advaat.dat};
\end{axis}
\end{tikzpicture}

  \end{minipage}%
}%
\begin{figure}[t]
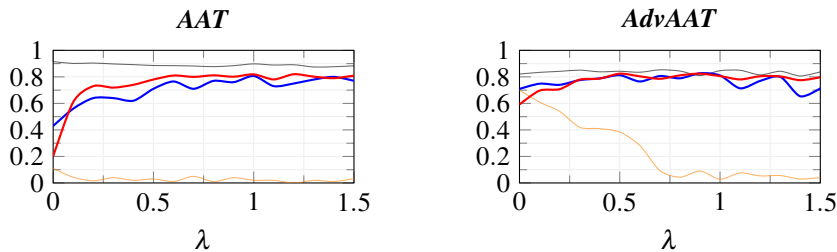

  \null\hfill %
  \begin{minipage}[t]{.43\linewidth}
    \input{plots/aat_lambda}%
  \end{minipage}%
 \hfill %
  \begin{minipage}[t]{.43\linewidth}
    \input{plots/advaat_lambda}%
  \end{minipage}%
 \hfill\null\par
  \caption{Estimated attributional robustness (\textcolor{blue}{$\huge{\bullet}$} \texttt{IN} and \textcolor{red}{$\huge{\bullet}$} \texttt{CO}), natural (\textcolor{black}{$\huge{\bullet}$} \texttt{NA}) and adversarial (\textcolor{orange}{$\huge{\bullet}$} \texttt{AA}) accuracies for our \textit{AAT} (left) and \textit{AdvAAT} (right) methods, evaluated on Fashion-MNIST varying the regularization parameter $\lambda$.}
  \label{fig:lambdarobustness}
\end{figure}
\begin{figure}[t]
\footnotesize%
\subfigure[]{
    \newcommand{\initpltwidth}{0.2}

\begin{tikzpicture}
\pgfplotsset{set layers}
\begin{axis}[
    width = \initpltwidth\textwidth,
    scale only axis,
    xmin=0, xmax=3,
    tick style={draw=none},
    xtick={0.5,1.5,2.5},
    xticklabels={Nat, Adv, \textit{AAT}},
    axis y line*=left,
    ylabel={\ref{plt:init_rob_topk} \texttt{IN}},
    ylabel style = {align=center},
    scatter/classes={%
		td={blue},
		tc={blue},
		tu={blue},
		tku={blue},
		tkn={blue},
		txu={blue},
		txn={blue},
        t={blue}
		},
    ]
    \addlegendimage{only marks, mark=oplus,black}
    \addlegendimage{only marks, mark=x,black}
    \addlegendimage{only marks, mark=square,black}
    \addlegendimage{only marks, mark=star,black}
    \addlegendimage{only marks, mark=triangle,black}
    \addlegendimage{only marks, mark=diamond,black}
    \addlegendimage{only marks, mark=pentagon,black}
    \addplot[scatter, only marks, scatter src=explicit symbolic]%
	table[meta=label] {
        x     y      label
        0.3   0.23   td
        0.3   0.09   tc
        0.3   0.18   tu
        0.3   0.13   tku
        0.3   0.10   tkn
        0.3   0.27   txu
        0.3   0.26   txn

        1.3   0.35   td
        1.3   0.21   tc
        1.3   0.40   tu
        1.3   0.12   tku
        1.3   0.11   tkn
        1.3   0.21   txu
        1.3   0.37   txn

        2.3   0.39   td
        2.3   0.30   tc
        2.3   0.33   tu
        2.3   0.38   tku
        2.3   0.36   tkn
        2.3   0.38   txu
        2.3   0.39   txn
	};
	\label{plt:init_rob_topk}
\end{axis}

\begin{axis}[
    width = \initpltwidth\textwidth,
    scale only axis,
    axis y line*=right,
    axis x line=none,
    xmin=0, xmax=3,
    ylabel={\ref{plt:init_rob_kendall} \texttt{CO}},
    ylabel style = {align=center},
    scatter/classes={%
		kd={red},
		kc={red},
		ku={red},
		kku={red},
		kkn={red},
		kxu={red},
		kxn={red},
		k={red}
		}
    ]
    \addplot[scatter, only marks, scatter src=explicit symbolic]%
	table[meta=label] {
        x     y      label
        0.6   0.20   kd
        0.6   0.03   kc
        0.6   0.13   ku
        0.6   0.08   kku
        0.6   0.06   kkn
        0.6   0.19   kxu
        0.6   0.20   kxn

        1.6   0.05   kd
        1.6   0.02   kc
        1.6   0.08   ku
        1.6   0.01   kku
        1.6   0.01   kkn
        1.6   0.45   kxu
        1.6   0.55   kxn

        2.6   0.28   kd
        2.6   0.18   kc
        2.6   0.24   ku
        2.6   0.24   kku
        2.6   0.24   kkn
        2.6   0.27   kxu
        2.6   0.28   kxn
	};
    \label{plt:init_rob_kendall}
\end{axis}

\end{tikzpicture}
    \label{fig:initialization_robustness}
}
\quad
\subfigure[]{
    \input{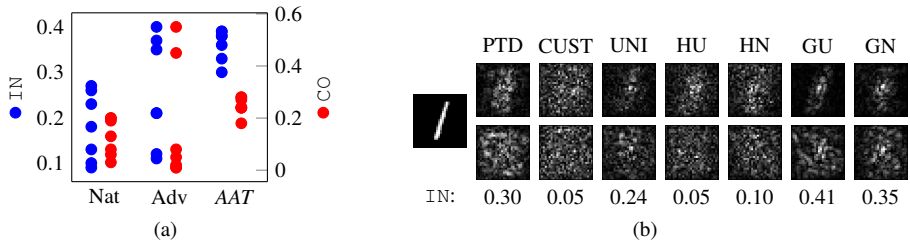}
    \label{fig:init_gradient_maps}
}%
\caption{(a) Estimated attributional robustness (\textcolor{blue}{$\huge{\bullet}$} \texttt{IN} and \textcolor{red}{$\huge{\bullet}$} \texttt{CO}) of SM (Equation \ref{eqn:saliencymap}) with \textit{seven} different initializations for the natural (Nat), adversarially (Adv) and attributionally (\textit{AAT}) trained models on MNIST. (b) Gradient maps (SM) and their attacked maps of the natural model trained on MNIST for different weight initializations.}
\label{fig:initialization}
\end{figure}
\paragraph{Dependency on the regularization parameter ($\lambda$).} We examine the influence of the regularization parameter $\lambda$ on the estimated robustness of attributions and predictions for our CNN trained on Fashion-MNIST. We chose this dataset because it is slightly more complex than MNIST, yet the computational burden of training is low. We train our \textit{AAT} and \textit{AdvAAT} models with $\lambda$-values varying from 0 to 1.5 and examine their robustness. Figure \ref{fig:lambdarobustness} contains the natural (\texttt{NA}) and adversarial (\texttt{AA}) accuracies as well as the attribution robustness metrics (\texttt{IN} and \texttt{CO}) for the \textit{AAT} models to the left and \textit{AdvAAT} models to the right. We observe that for both methods, higher $\lambda$ values result in increased AR, with saturation occurring at values above 1. Moreover, for \textit{AdvAAT}, the adversarial accuracy drops with increasing $\lambda$, while AR metrics increase, controlling the trade-off between adversarial and attribution robustness.%
%
%
\paragraph{Dependency on network parameter initialization.} Our experiments have shown that gradient-based attribution maps and their robustness estimates can depend on the initialization of the weights in the network. While resulting in nearly identical natural and adversarial accuracies, differently initialized networks yield considerably different robustness of gradient maps. We exemplify this with our natural, adversarially and \textit{AAT} trained models on MNIST, by reporting their corresponding performance and attribution robustness estimates for seven different network weight initializations. These are the default PyTorch \cite{pytorch} initialization (PTD), a custom initialization taken from \cite{igsum} (CUST), a random uniform initialization of weights (UNI) as well as the default PyTorch He \cite{kaiming} and Glorot \cite{xavier} uniform and normal (HU, HN, GU and GN) initializations (as listed in Figure \ref{fig:init_gradient_maps} from left to right). Figure \ref{fig:initialization_robustness} reports the resulting attributional robustness estimates for the initialization methods. Both for natural and adversarially robust models, the variance of \texttt{IN} and \texttt{CO} is significant across the different initializations. We expect this behaviour, as heuristic search algorithms like SGD depends strongly on initial conditions. The gradient maps look notably different as well, as reported in Figure \ref{fig:init_gradient_maps}. This dependency is partly mitigated by our \textit{AAT} method, but still present.%
%
%
%
\paragraph{Dependency on the tightness parameter of the ReLU approximation  ($\beta$).}%
\begin{wrapfigure}{R}{0.5\linewidth}
\footnotesize%
\begin{tikzpicture}
\begin{axis}[
xmin = 0.1, xmax = 100.0,
xmode=log,
ymin = 0.15, ymax = 0.4,
grid = both,
minor tick num = 1,
major grid style = {lightgray!25},
minor grid style = {lightgray!25},
width = \linewidth,
height = 0.45\linewidth,
xlabel = {$\beta$},
ylabel = {\texttt{IN}},
]
\addplot[
smooth,
thick,
red,
] file[skip first] {data/beta_robust.dat};
\end{axis}
\end{tikzpicture}

\caption{Estimated attributional robustness (\texttt{IN}) of SM for the natural model trained on MNIST, varying the $\beta$ parameter of the ReLU aprroximation.}
\label{fig:beta_robustness}
\end{wrapfigure}
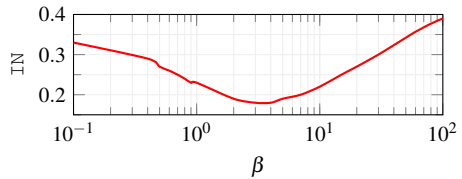%
Figure \ref{fig:beta_robustness} shows the estimated \textit{Top-K intersection} of the natural MNIST model while using different $\beta$ values for the second gradient approximation. We observe that by varying this parameter, the \textit{Top-K intersection} changes considerably. We further observe that by setting $\beta$ too extreme, second gradients vanish, resulting in the IFIA attack not being able to find good adversarial inputs. Previous work \cite{geometryblame} has already shown the dependency of AR on $\beta$, however, we are the first to only use this approximation for the second order gradients. Therefore, we keep saliency maps unchanged, giving a better estimate for the true attribution robustness of ReLU networks.%

\section{Conclusion and Future Work}
This work introduced a generalized notion of attributional robustness with FAR providing objectives for increasing the robustness of explanations in DNNs. This allows direct optimization for robust attributions, with optionally coupling it to robust predictions. We showed how current existing objectives can be instantiated from this framework. Moreover, we provided novel instantiations of FAR, \textit{AAT} and \textit{AdvAAT}, which directly optimize for high correlation of attributions as well as robust predictions for similar inputs. They perform comparably to or better than current state of the art methods in terms of AR, utilizing fewer assumptions and generalizing better (see Section \ref{sec:experiments}). Finally, we identified parameter dependencies of robust attributions that necessitate careful assessment of methods on their dependencies on these parameters.\\
{\indent}This work opens up many interesting directions for future research. First, we are interested in assessing the robustness of \textit{non-differentiable maps} like Occlusion \cite{occlusion}, utilizing gradient estimation techniques. Second, exploring the \textit{connection between robust predictions and robust attributions} might lead to additional insights into the decision process of neural networks, enhancing interpretability. Lastly, since second order gradients in DNNs seem highly irregular, and their optimization is hard, the assessment of \textit{additional training parameters} that influence AR would help further research in establishing fair comparison and tracking of true progress in this area.

\bibliography{main.bib}

\newpage
\appendix
\section{Supplementary material}
\subsection{Proofs}
\setcounter{page}{1}
In this section of the appendix, we show how current robust training methods can be derived from our FAR objectives. The notation is kept as in Chapter \ref{sec:far} from the paper.%
\subsubsection{Proof of Equation (\ref{eqn:madryrobust})}
Setting $\lambda = 0$ in Equation (\ref{eqn:robusttraining}) straightforwardly results as follows.%
\begin{equation*}
\begin{split}
    \theta^* & {}={} \underset{\theta}{\arg\min} \,  \sum_{\mathbf{x} \in \mathrm{D}} \,  \max_{\|\mathbf{\tilde x} - \mathbf{x}\|_{\mathrm{p}} < \varepsilon} \, \big\{\, l(\mathbf{\tilde x}, y, f)
    + \lambda \, \cdot \, d_{\mathrm{s}} \big[\, \mathrm{S}(\mathbf{\tilde{x}}, f), \mathrm{S}^T(\mathbf{x}, f)\, \big] \, \big\}\\%
    & {}={} \underset{\theta}{\arg\min} \,  \sum_{\mathbf{x} \in \mathrm{D}} \,  \max_{\|\mathbf{\tilde x} - \mathbf{x}\|_{\mathrm{p}} < \varepsilon} \, \big\{\, l(\mathbf{\tilde x}, y, f)
    + 0 \, \cdot \, d_{\mathrm{s}} \big[\, \mathrm{S}(\mathbf{\tilde{x}}, f), \mathrm{S}^T(\mathbf{x}, f)\, \big] \, \big\}\\%
    & {}={} \underset{\theta}{\arg\min} \,  \sum_{\mathbf{x} \in \mathrm{D}} \,  \max_{\|\mathbf{\tilde x} - \mathbf{x}\|_{\mathrm{p}} < \varepsilon}\, l(\mathbf{\tilde x}, y, f)
\end{split}
\end{equation*}%
\subsubsection{Proof of Equation (\ref{eqn:ignormequation})}%
In order to derive the IG-NORM objective from Equation (\ref{eqn:robustregularizationtraining}), we use IG as saliency maps, set the distance metric $d_{\mathrm{s}}$ to be the $\mathrm{L}_1$-norm induced distance and choose the baseline for IG to be the unperturbed input to the network, i.e. $\mathbf{b} {}={} \mathbf{x}$. Then, Equation (\ref{eqn:robustregularizationtraining}) becomes as follows.
\begin{equation*}
\begin{split}
    \theta^* & {}={} \underset{\theta}{\arg\min} \,  \sum_{\mathbf{x} \in \mathrm{D}} \, \big\{\, l(\mathbf{x}, y, f) \,
    + \lambda \, \cdot \, \max_{\|\mathbf{\tilde x} - \mathbf{x}\|_{\mathrm{p}} < \varepsilon} \, d_{\mathrm{s}} \big[\, \mathrm{S}(\mathbf{\tilde{x}}, f), \mathrm{S}^T(\mathbf{x}, f)\, \big] \; \big\}\\%
    & {}={} \underset{\theta}{\arg\min} \,  \sum_{\mathbf{x} \in \mathrm{D}} \, \big\{\, l(\mathbf{x}, y, f) \,
    + \lambda \, \cdot \, \max_{\|\mathbf{\tilde x} - \mathbf{x}\|_{\mathrm{p}} < \varepsilon} \, \| \mathrm{S}(\mathbf{\tilde{x}}, f) {}-{} \mathrm{S}^T(\mathbf{x}, f)\, \|_1 \big\}\\%
    & {}={} \underset{\theta}{\arg\min} \,  \sum_{\mathbf{x} \in \mathrm{D}} \, \big\{\, l(\mathbf{x}, y, f) \,
    + \lambda \, \cdot \, \max_{\|\mathbf{\tilde x} - \mathbf{x}\|_{\mathrm{p}} < \varepsilon} \, \| \mathrm{IG}(\mathbf{\tilde{x}}, \mathbf{x}) {}-{} \mathrm{IG}(\mathbf{x}, \mathbf{x})\, \|_1 \big\}\\%
    & {}={} \underset{\theta}{\arg\min} \,  \sum_{\mathbf{x} \in \mathrm{D}} \, \big\{\, l(\mathbf{x}, y, f) \,
    + \lambda \, \cdot \, \max_{\|\mathbf{\tilde x} - \mathbf{x}\|_{\mathrm{p}} < \varepsilon} \, \| \mathrm{IG}(\mathbf{\tilde{x}}, \mathbf{x})\, \|_1 \big\}\\%
\end{split}
\end{equation*}%
Note that $\mathrm{IG}(\mathbf{x}, \; \mathbf{x}) {}={} 0$ holds due to the completeness axiom of IG. The IG-SUM-NORM objective can analogously be derived from Equation (\ref{eqn:robusttraining}).%
\subsubsection{Proof of Equation (\ref{eqn:alignmentobjective})}%
The training regularization of the Align method considers the scalar product between input gradients and the original input image. To derive their objective from our framework, we have to set S to the expression given in Equation (\ref{eqn:alignmentobjective}), with the dissimilarity $d_{\mathrm{s}}(\mathbf{x}, \mathbf{y}) {}={} \log \, \big\{ 1 + \exp \, \big[-{} \sum_{i\in\mathrm{dim}(\mathbf{x})}(x_i {}-{} y_i) \, \big] \, \big\}$ and $\mathrm{S}^T = \mathbf{0}$. Equation (\ref{eqn:robustregularizationtraining}) then becomes as follows.%
\begin{equation*}
\begin{split}
    \theta^* & {}={} \underset{\theta}{\arg\min} \,  \sum_{\mathbf{x} \in \mathrm{D}} \, \big\{\, l(\mathbf{x}, y, f) \,
    + \lambda \, \cdot \, \max_{\|\mathbf{\tilde x} - \mathbf{x}\|_{\mathrm{p}} < \varepsilon} \, d_{\mathrm{s}} \big[\, \mathrm{S}(\mathbf{\tilde{x}}, f), \mathrm{S}^T(\mathbf{x}, f)\, \big] \; \big\}\\%
    & {}={} \underset{\theta}{\arg\min} \,  \sum_{\mathbf{x} \in \mathrm{D}} \, \big\{\, l(\mathbf{x}, y, f) \,
    + \lambda \, \cdot \, \max_{\|\mathbf{\tilde x} - \mathbf{x}\|_{\mathrm{p}} < \varepsilon} \, d_{\mathrm{s}} \big[\, \mathrm{cos} \big[ {g^y_\mathbf{\tilde{x}}(\mathbf{\tilde{x}})}, \mathbf{x} \big] \,
    -{} \, \mathrm{cos} \big[ {g^{\bar{y}}_\mathbf{\tilde{x}}(\mathbf{\tilde{x}})}, \mathbf{x} \big]\; \big\}\\%
    & {}={} \underset{\theta}{\arg\min} \,  \sum_{\mathbf{x} \in \mathrm{D}} \, \big\{\, l(\mathbf{x}, y, f) \,
    + \lambda \, \cdot \, \max_{\|\mathbf{\tilde x} - \mathbf{x}\|_{\mathrm{p}} < \varepsilon} \, d_{\mathrm{s}} \big[\, \mathrm{cos} \big[ {g^y_\mathbf{\tilde{x}}(\mathbf{\tilde{x}})}, \mathbf{x} \big] \,
    -{} \, \mathrm{cos} \big[ {g^{\bar{y}}_\mathbf{\tilde{x}}(\mathbf{\tilde{x}})}, \mathbf{x} \big]\; \big\}\\%
    & {}={} \underset{\theta}{\arg\min} \,  \sum_{\mathbf{x} \in \mathrm{D}} \, \big\{\, l(\mathbf{x}, y, f) \,
    + \lambda \, \cdot \, \max_{\|\mathbf{\tilde x} - \mathbf{x}\|_{\mathrm{p}} < \varepsilon} \, \log \; \{\; 1 + \exp \big[\,  \,
    \mathrm{cos} [ {g^{\bar{y}}_\mathbf{\tilde{x}}(\mathbf{\tilde{x}})}, \mathbf{x} ]\;  -{} \mathrm{cos} [ {g^y_\mathbf{\tilde{x}}(\mathbf{\tilde{x}})}, \mathbf{x} ]\big] \;\} \; \big\}\\%
\end{split}
\end{equation*}%
%
\subsection{Parameters and architectures}%
\begin{table}[h]
\resizebox{\textwidth}{!}{%
\begin{tabular}{c|c|c|c|c|c|c}
\toprule
\multicolumn{2}{c|}{\textit{Dataset}}                                                                             & MNIST            & Fashion-MNIST    & CIFAR-10             & GTSRB               & Restr. Imagenet   \\ \hline\hline
\multicolumn{2}{c|}{\textit{Architecture}}                                                                        & CNN \cite{lenet} & CNN \cite{lenet} & ResNet \cite{resnet} & ResNet \cite{resnet} & ResNet \cite{resnet} \\ \hline
\multirow{3}{*}{\texttt{AA}}     & \textit{Attack}             & \multicolumn{5}{c}{PGD}                                                                                  \\
                                                                                    & \textit{Steps}              & \multicolumn{5}{c}{40}                                                                                   \\
                                                                                    & \textit{Rel. stepsize}  & \multicolumn{5}{c}{0.03}                                                                                 \\ \hline
\multirow{7}{*}{\texttt{AR}} & \textit{Attack}             & \multicolumn{5}{c}{IFIA}                                                                                 \\
                                                                                    & \textit{Explainer}          & \multicolumn{5}{c}{Integrated Gradients with baseline 0}                                                 \\
                                                                                    & $d_s$      & \multicolumn{5}{c}{Sum-Top-K}                                                                            \\
                                                                                    & \textit{Steps}              & \multicolumn{5}{c}{7}                                                                                    \\
                                                                                    & \textit{Rel. stepsize}  & \multicolumn{5}{c}{1.2/7}                                                                                \\
                                                                                    & $\beta$               & \multicolumn{5}{c}{1.0}                                                                                  \\ \cline{3-7}
                                                                                    & \textit{k}                  & 50    & 50            & 100      & 100    & 300                                                          \\ \hline
\multicolumn{2}{c|}{$\varepsilon$}                                                                             & 0.3   & 0.1           & 0.03     & 0.03   & 0.01                                                         \\ \hline
\multicolumn{2}{c|}{\textit{Number of restarts}}                                                                  & \multicolumn{5}{c}{3}                                                                                    \\
\hline              
\end{tabular}%
}
\caption{\footnotesize{Evaluation parameters}}
\label{tbl:evalparams}
\end{table}
\begin{table}[h!]
\resizebox{\textwidth}{!}{%
\begin{tabular}{c|c|c|c|c|c|c}
\toprule
\multicolumn{2}{c|}{\textit{Dataset}}                          & MNIST                      & Fashion-MNIST              & CIFAR-10                   & GTSRB                      & Restr. Imagenet        \\ \hline \hline
\multirow{4}{*}{\texttt{Nat}}    & Optimizer                           & \multicolumn{5}{c}{Adam}                                                                                                                       \\
                        & \textit{Epochs}                     & \multicolumn{5}{c}{50}                                                                                                                         \\ \cline{3-7}
                        & \textit{Batch size}                 & 50                         & 50                         & 128                        & 128                        & 32                         \\
                        & \textit{LR}                         & 0.001                      & 0.001                      & 0.01                       & 0.01                       & 0.01                       \\ \hline
\multirow{5}{*}{\texttt{Adv}}    & \textit{Optimizer}                  & \multicolumn{5}{c}{Adam}                                                                                                                       \\
                        & \textit{Epochs}                     & \multicolumn{5}{c}{50}                                                                                                                         \\ \cline{3-7}
                        & \textit{Batch size}                 & 50                         & 50                         & 128                        & 128                        & 32                         \\
                        & \textit{LR}                         & 0.0001                     & 0.001                      & 0.001                      & 0.001                      & 0.001                      \\ \cline{3-7}
                        & \textit{Adv. ratio}                 & \multicolumn{5}{c}{0.7}                                                                                                                        \\ \hline
\multirow{5}{*}{\texttt{Align}}  & \textit{Optimizer}                  & \multicolumn{5}{c}{Adam}                                                                                                                       \\
                        & \textit{Epochs}                     & \multicolumn{5}{c}{50}                                                                                                                         \\ \cline{3-7}
                        & \textit{Batch size}                 & 50                         & 50                         & -                          & -                          & 32                         \\
                        & \textit{LR}                         & 0.0001                     & 0.0001                     & -                          & -                          & 0.0001                     \\
                        & $\lambda$                     & 0.5                        & 0.5                        & -                          & -                          & 0.5                        \\ \hline
\multirow{5}{*}{\texttt{AAT}}    & \textit{Optimizer}                  & \multicolumn{5}{c}{Adam}                                                                                                                       \\
                        & \textit{Epochs}                     & \multicolumn{5}{c}{50}                                                                                                                         \\ \cline{3-7}
                        & \textit{Batch size}                 & 50                         & 50                         & 128                        & 128                        & 32                         \\
                        & \textit{LR}                         & 0.0001                     & 0.0001                     & 0.0001                     & 0.0001                     & 0.0001                     \\
                        & $\lambda$                     & 0.5                        & 1.0                        & 2.0                        & 0.5                        & 1.5                        \\ \hline
\multirow{5}{*}{\texttt{AdvAAT}} & \textit{Optimizer}                  & \multicolumn{5}{c}{Adam}                                                                                                                       \\
                        & \textit{Epochs}                     & \multicolumn{5}{c}{50}                                                                                                                         \\ \cline{3-7}
                        & \textit{Batch size}                 & 50                         & 50                         & 128                        & 128                        & 32                         \\
                        & \textit{LR}                         & 0.0001                     & 0.0001                     & 0.0001                     & 0.0001                     & 0.0001 \\
                        & $\lambda$                     & 0.5                        & 0.5                        & 0.5                        & 0.2                        & 0.5  \\ \hline
\end{tabular}%
}
\caption{Training parameters}
\label{tbl:trainparams}
\end{table}
We conduct experiments on five vision datasets (MNIST, Fashion-MNIST, CIFAR-10, GTSRB and Restricted Imagenet) to compare our attributional robustness method to state of the art algorithms. Each model is implemented in PyTorch v1.3.1 and is trained distributedly on six NVIDIA Tesla V100 GPUs with the PyTorch Distributed Data Parallel wrapper. We fix all seeds to 42. Table \ref{tbl:evalparams} contains the evaluation parameters of our experiments, Table \ref{tbl:trainparams} the training parameters. We finetune the natural model to train our robust methods. If we do not mention a specific parameter, it is set to the default value in PyTorch v1.3.1. Moreover, the parameters values of IFIA during training are kept as the values during evaluation.
\subsection{Initialization methods}
\begin{wraptable}{r}{0.65\textwidth}%
\vspace{-10pt}
\begin{tabular}{cccccc}
\toprule
\texttt{Init.}        & \texttt{Model} & \texttt{NA} & \texttt{AA} & \texttt{IN} & \texttt{CO} \\ \hline\hline
\multirow{3}{*}{PTD}  & Nat   & 99.1\%      & 0.0\%       & 0.23        & 0.20        \\ 
                      & Adv   & 99.0\%      & 93.9\%      & 0.35        & 0.05        \\ 
                      & \textit{AAT}  & 98.9\%      & 8.7\%       & 0.39        & 0.28        \\ \hline
\multirow{3}{*}{CUST} & Nat   & 98.8\%      & 0.0\%       & 0.09        & 0.03        \\ 
                      & Adv   & 98.8\%      & 88.9\%      & 0.21        & 0.02        \\ 
                      & \textit{AAT}  & 98.6\%      & 8.7\%       & 0.30        & 0.18        \\ \hline
\multirow{3}{*}{UNI}  & Nat   & 99.2\%      & 0.0\%       & 0.18        & 0.13        \\ 
                      & Adv   & 98.9\%      & 93.6\%      & 0.40        & 0.08        \\ 
                      & \textit{AAT}  & 98.7\%      & 5.5\%       & 0.33        & 0.24        \\ \hline
\multirow{3}{*}{HU}   & Nat   & 99.2\%      & 0.0\%       & 0.13        & 0.08        \\ 
                      & Adv   & 99.0\%      & 93.6\%      & 0.12        & 0.01        \\ 
                      & \textit{AAT}  & 98.3\%      & 7.2\%       & 0.38        & 0.24        \\ \hline
\multirow{3}{*}{HN}   & Nat   & 99.2\%      & 0.0\%       & 0.10        & 0.06        \\ 
                      & Adv   & 99.1\%      & 93.6\%      & 0.11        & 0.01        \\ 
                      & \textit{AAT}  & 98.5\%      & 4.3\%       & 0.36        & 0.24        \\ \hline
\multirow{3}{*}{GU}   & Nat   & 99.2\%      & 0.0\%       & 0.27        & 0.19        \\ 
                      & Adv   & 99.0\%      & 93.6\%      & 0.21        & 0.45        \\ 
                      & \textit{AAT}  & 98.8\%      & 6.4\%       & 0.38        & 0.27        \\ \hline
\multirow{3}{*}{GN}   & Nat   & 99.2\%      & 0.0\%       & 0.26        & 0.20        \\ 
                      & Adv   & 99.0\%      & 94.0\%      & 0.37        & 0.55        \\ 
                      & \textit{AAT}  & 98.7\%      & 9.1\%       & 0.39        & 0.28        \\ \hline

\end{tabular}
\caption{Estimated attributional robustness (\texttt{IN} and \texttt{CO}) for several different initialization methods (\texttt{Init.}). The results are reported for models trained naturally (Nat), adversarially (Adv) as well as with our \textit{AAT} objective on MNIST. The natural and adversarial accuracy is given in the \texttt{NA} and \texttt{AA} columns. While accuracies of the models are similar, their estimated attributional robustness varies significantly throughout the initializations.}\label{tbl:initrobustness}%
\end{wraptable}%
We use seven different initialization methods for addressing the dependency of attributional robustness on the initialization. These are detailed in the next paragraphs. If a parameter is not mentioned, it is kept as the default value defined in PyTorch. The training setup is kept constant for each initialization, and corresponds to the setup mentioned in the previous section for the different models.\\%
{\textbf{PTD.}} Default PyTorch initialization for linear and convolutional layers. This is the He uniform initialization with $a=\sqrt{5}$ for the weights and a uniform initialization with bounds $\pm b {}={} \pm 1/\sqrt{\mathtt{fan\_in}}$ for the bias terms.\\
{\textbf{CUST.}} Custom initialization method. Weights are initialized utilizing a zero-centered normal distribution with a standard deviation of 0.1, and biases are initialized to be 0.1, both for linear and convolutional layers.\\
{\textbf{UNI.}} Uniform initialization method. Weights and biases are initialized utilizing a uniform distribution with bounds $\pm b {}={} \pm 0.1$ for all layers.\\
{\textbf{HU.}} He uniform initialization method. Weights are initialized utilizing the default PyTorch He uniform initialization, biases are set to zero.\\
{\textbf{HN.}} He uniform initialization method. Weights are initialized utilizing the default PyTorch He normal initialization, biases are set to zero.\\
{\textbf{GU.}} Glorot uniform initialization method. Weights are initialized utilizing the default PyTorch Glorot uniform initialization, biases are set to zero.\\
{\textbf{GN.}} Glorot normal initialization method. Weights are initialized utilizing the default PyTorch Glorot normal initialization, biases are set to zero.
\end{document}